\documentclass{article}
\PassOptionsToPackage{numbers}{natbib}
\usepackage[final]{neurips_2026}

\makeatletter
\providecommand{\@trackname}{}
\makeatother
\usepackage{amsmath,amssymb,amsfonts,amsthm,mathtools}
\usepackage[utf8]{inputenc}
\usepackage[T1]{fontenc}
\usepackage{times}
\usepackage{hyperref}
\usepackage{url}
\usepackage{booktabs}
\usepackage{nicefrac}
\usepackage{microtype}
\usepackage{xcolor}
\usepackage{colortbl}
\usepackage{graphicx}
\usepackage{subcaption}
\usepackage{multirow}
\usepackage{algorithm}
\usepackage{algorithmic}
\usepackage{enumitem}
\usepackage{wrapfig}
\usepackage[most]{tcolorbox}
\newtcolorbox{keyinsight}{
    enhanced,
    colback=blue!4,
    colframe=blue!55!black,
    boxrule=0.6pt,
    arc=3pt,
    left=8pt, right=8pt, top=5pt, bottom=5pt,
    before skip=8pt, after skip=10pt,
    fontupper=\small,
    before upper={\textbf{\color{blue!55!black}Key insight.}~}
}
\usepackage[capitalize,noabbrev]{cleveref}
\usepackage{aliascnt}
\theoremstyle{plain}

\theoremstyle{definition}
\newaliascnt{definition}{theorem}

\aliascntresetthe{definition}
\newaliascnt{assumption}{theorem}

\aliascntresetthe{assumption}
\theoremstyle{remark}
\newaliascnt{remark}{theorem}

\aliascntresetthe{remark}
\crefname{assumption}{Assumption}{Assumptions}
\Crefname{assumption}{Assumption}{Assumptions}
\crefname{definition}{Definition}{Definitions}
\Crefname{definition}{Definition}{Definitions}
\crefname{remark}{Remark}{Remarks}
\Crefname{remark}{Remark}{Remarks}
\newcommand{\KL}{\mathrm{KL}}

\newcommand{\cL}{\mathcal{L}}

\newcommand{\teacher}{p_{\mathrm{T}}}        % teacher conditioned on v (with fine-grained visual detail)
\newcommand{\teachert}{\tilde{p}_{\mathrm{T}}}  % teacher conditioned on tilde-v (without fine-grained visual detail)
\newcommand{\student}{p_{\mathrm{S}}}        % student (with image)
\newcommand{\va}{a}                           % per-token VA symbol
\newcommand{\vabar}{\bar{a}}                  % trajectory-averaged VA

\newcommand{\Lpgd}{\cL_{\text{VA-OPD}}}       % total VA-OPD loss

\newcommand{\tup}[1]{{\scriptsize\textcolor{teal}{$\uparrow$#1}}}
\title{Visual-Advantage On-Policy Distillation\\for Vision-Language Models}

\author{
\textbf{Ruiqi Liu}$^{1,2*}$\quad
\textbf{Xiaolei Lv}$^{3*}$\quad
\textbf{Gengsheng Li}$^{1}$\quad
\textbf{Ximo Zhu}$^{3}$\quad
\textbf{Zhiheng Wang}$^{4}$\\
\textbf{Zhengbo Zhang}$^{1}$\quad
\textbf{Junkai Chen}$^{1}$\quad
\textbf{Zhiheng Li}$^{1}$\quad
\textbf{Bo Li}$^{3}$\quad
\textbf{Jun Gao}$^{3}$\quad
\textbf{Shu Wu}$^{1\dagger}$\\[6pt]
$^{1}$Institute of Automation, CAS\quad
$^{2}$School of Advanced Interdisciplinary Sciences, UCAS\\
$^{3}$Hello Group Inc.\quad
$^{4}$Sun Yat-sen University\quad
$^{*}$Equal contribution\quad
}

\begin{document}
\maketitle

\begin{abstract}
On-policy knowledge distillation has proven effective for language models, yet its application to vision-language models (VLMs) remains underexplored. We observe that standard on-policy distillation can improve a student's output quality while failing to strengthen its reliance on visual input: on vision-critical tokens, the student's predictions remain largely unchanged whether or not fine-grained visual detail is present, even though the teacher's predictions depend heavily on it. To make this difference observable, we introduce \emph{visual advantage} (VA), the token-level log-probability difference when the teacher scores a student-generated rollout with versus without access to fine-grained visual detail. VA is concentrated in a small minority of tokens, and these high-VA tokens are the ones that actually carry the visual supervision signal. This motivates a distillation objective that treats them differently from language scaffolding, so their contribution is not diluted by the abundant surrounding language tokens. We propose \textbf{Visual-Advantage On-Policy Distillation (VA-OPD)}, which uses VA at two granularities: rollout-level reweighting by trajectory-averaged VA, and token-level KL averaged within high-VA and low-VA groups separately. We train on two math datasets (Geometry3K and ViRL39K) and evaluate on eight benchmarks covering both mathematical reasoning and visual understanding, across three teacher sizes (4B, 8B, and 32B) on the Qwen3-VL family. VA-OPD improves over standard on-policy distillation on every benchmark, with the gain growing monotonically along both the teacher-size and data-scale axes, suggesting that these factors compound consistently.

\end{abstract}
\section{Introduction}
\label{sec:intro}

\begin{figure}[t]
    \centering
    \includegraphics[width=\textwidth]{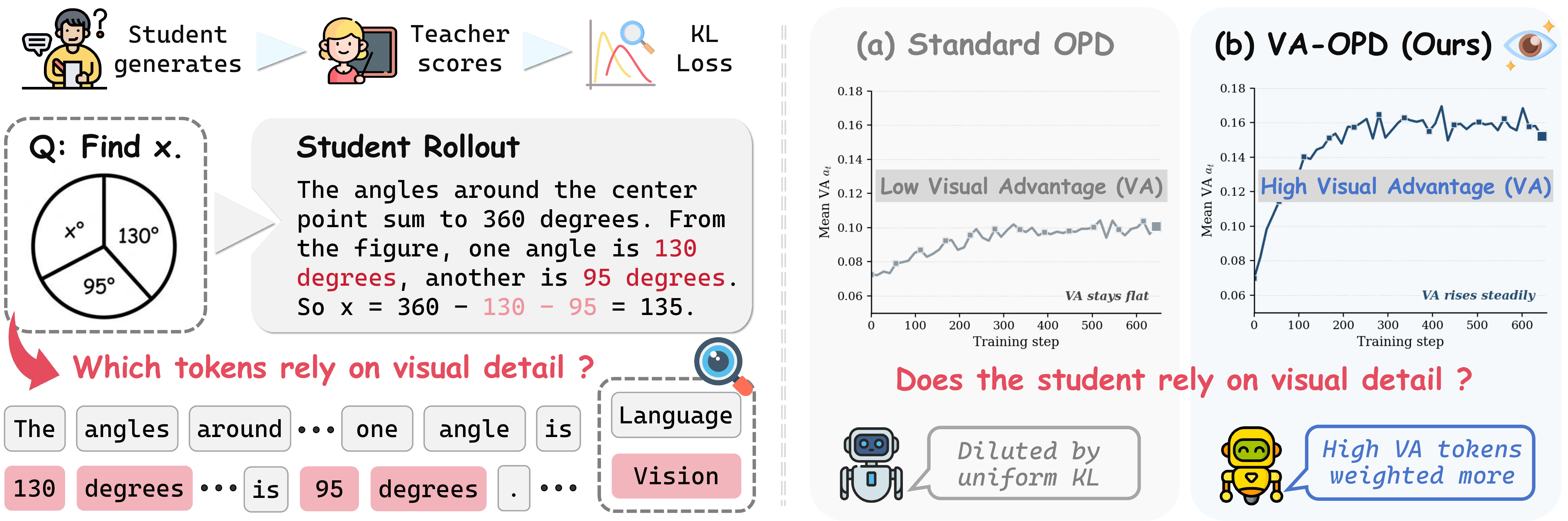}
    \caption{\textbf{Token-level visual reliance in VLM distillation.}
    \emph{Left:} in a student rollout, only a minority of tokens (red) depend on
    the image, while the rest are language-template tokens (gray). Standard OPD
    applies a uniform KL weight to all tokens, diluting the learning signal on
    vision-critical positions. \emph{Right:} teacher-scored visual advantage (VA),
    averaged over the student's rollouts, tracked across training, for two
    distillation methods.
    Under \textbf{Standard OPD}, student-rollout VA stays nearly flat: matching
    the teacher's outputs does not produce rollouts whose tokens actually
    require the image. Under \textbf{VA-OPD}, student-rollout VA rises steadily across training, indicating that the student increasingly relies on visual detail when generating image-dependent predictions.}
    \label{fig:hero}
    \vspace{-3mm}
\end{figure}

On-policy distillation has emerged as an effective strategy for compressing large language models~\citep{agarwal2024policy,gu2024minillm,yang2026learning}: a student learns by matching a teacher's output distribution on its own rollouts. Extending it to vision-language models (VLMs), however, surfaces a mismatch that text-only distillation never had to confront. Not all tokens in a VLM's response are created equal with respect to the image. As illustrated in \Cref{fig:hero} (left), in a typical VLM rollout the vast majority of tokens serve as language scaffolding or reasoning templates, and only a small minority are determined by visual evidence. Standard KL-based distillation weights every token uniformly, raising a basic question: does on-policy distillation actually push the student to rely on visual input, or is matching the teacher on the language scaffolding alone enough?

Answering this requires a token-level signal of visual dependency that a uniform KL cannot supply on its own. We introduce a simple counterfactual probe: for each token in a student-generated rollout, we score it under the teacher with and without fine-grained visual detail, and take the log-probability difference as the per-token \emph{visual advantage} (VA). VA quantifies, for each student-generated token, how much the teacher judges that predicting it should depend on fine-grained visual detail. Tracking VA on student rollouts during training reveals a concerning pattern (\Cref{fig:hero}, right): standard on-policy distillation steadily improves task accuracy, yet rollout VA stays nearly flat. The student's outputs resemble the teacher's, but their underlying visual reliance does not follow.

Two observations on student rollouts help explain this. VA is heavy-tailed: the top 10\% of tokens carry most of the total VA mass while the rest sit close to zero. A controlled mask ablation shows that this small minority is exactly where the visual supervision lives, since removing the KL loss on the top-VA tokens collapses accuracy while removing the same fraction of low-VA or random tokens has essentially no effect. Standard KL averages uniformly over all tokens, so the gradient contribution from the few high-VA tokens is diluted by the surrounding linguistic scaffolding.

We propose \textbf{Visual-Advantage On-Policy Distillation (VA-OPD)}, which uses VA at two levels of granularity to address this dilution. At the rollout level, the $K$ sibling rollouts under the same prompt vary in how much their tokens depend on fine-grained visual detail; we soft-weight each rollout by its trajectory-averaged VA relative to its siblings, so that rollouts with stronger visual reliance contribute more to the gradient. At the token level, we sort each rollout's tokens by VA, assign the top fraction and remainder to high- and low-VA groups, then separately average per-token KL per group before summing. Beyond Standard OPD, VA-OPD only adds one teacher forward pass on the image with fine-grained visual detail destroyed; it needs no extra annotations, reward models, or rollouts.

We validate VA-OPD across three teacher sizes (4B, 8B, and 32B), two training corpora (Geometry3K and ViRL39K), and eight benchmarks. VA-OPD improves over standard on-policy distillation on every benchmark. Gains are sizable throughout the math-reasoning benchmarks, and within visual understanding the largest gains come on those most reliant on fine-grained visual detail: HallusionBench $+2.5$ and AI2D $+2.4$, precisely the tasks that most require extracting the correct visual evidence. The gain amplifies under the larger ViRL39K corpus (Math Avg $+3.8$, Visual Avg $+2.5$). During VA-OPD training, VA on student rollouts rises alongside accuracy, indicating that the performance gain comes with a genuine strengthening of the student's reliance on visual detail during generation, rather than merely closer surface-level mimicry of the teacher's outputs.

Our contributions are:
\begin{itemize}[nosep,leftmargin=*]
    \item We find that in standard on-policy distillation for VLMs, since only a small minority of tokens depend on fine-grained visual detail, uniform-weighted KL dilutes the gradient on these critical tokens with the surrounding language scaffolding. As a result, the student aligns with the teacher's outputs but does not strengthen its reliance on fine-grained visual detail.
    \item We introduce \emph{visual advantage} (VA), which characterizes the degree of visual dependence at each token in student-generated rollouts. Based on this, we propose VA-OPD, which applies VA at two granularities: rollout-level reweighting by trajectory-averaged VA, and token-level KL averaged within high-VA and low-VA groups separately.
    \item Across three teacher sizes (4B, 8B, and 32B), two training corpora (Geometry3K and ViRL39K), and eight benchmarks, VA-OPD improves over standard on-policy distillation on every benchmark. The teacher-scored VA on student rollouts also rises with accuracy during VA-OPD training, showing that the gain comes with stronger visual reliance, not just closer output mimicry.
\end{itemize}

\section{Background}
\label{sec:prelim_sec}

We consider on-policy distillation of vision-language models: a teacher $\teacher$ and student $\student$, both conditioned on an image $v$ and query $q$. The student generates $K$ rollouts $\{y^{(k)}\}_{k=1}^K \sim \student(\cdot \mid v, q)$, and the teacher provides token-level targets via the reverse KL \citep{gu2024minillm,ko2024distillm}:
\begin{equation}
    \cL_{\mathrm{KL}} = \frac{1}{T}\sum_{t=1}^{T} \KL\!\left(\student(\cdot \mid v, q, y_{<t}) \;\|\; \teacher(\cdot \mid v, q, y_{<t})\right).
    \label{eq:standard_kl}
\end{equation}
We refer to training with this objective throughout the paper as \emph{Standard OPD}. The objective constrains which tokens the student predicts, not whether it attends to fine-grained visual detail: a student can minimize KL by exploiting language priors without extracting information from the diagram. We now introduce a diagnostic that makes this distinction observable (\Cref{sec:va}), then present empirical evidence that motivates our method (\Cref{sec:motivation}).

\subsection{Visual Advantage}
\label{sec:va}

For each token in a student rollout, we want to know how much the teacher relies on fine-grained visual detail to predict it. The probe is simple: score the same token under the teacher in two matched conditions, with and without fine-grained visual detail, and take the non-negatively rectified log-probability difference as the per-token \emph{visual advantage} (VA):
\begin{equation}
    \va_t = \max\!\left(\log \teacher(y_t \mid v, q, y_{<t}) - \log \teacher(y_t \mid \tilde{v}, q, y_{<t}),\; 0\right),
    \label{eq:va}
\end{equation}
where $\tilde{v}$ is a degraded version of $v$ with fine-grained visual detail destroyed. A high $\va_t$ means the teacher cannot predict the token without fine-grained visual detail, e.g., a precise number that must be read off the diagram; $\va_t \approx 0$ means that the token is language scaffolding or relies only on robust global image information that survives degradation. Thus, VA is the teacher's per-token judgment of whether predicting a token depends on fine-grained visual detail. Zero rectification discards positions where the logprob under $\tilde{v}$ exceeds that under $v$, which mostly reflect low-confidence noise.

We realize $\tilde{v}$ by downsampling the image to 10\% of its spatial resolution with bilinear interpolation and upsampling back via nearest-neighbor. This destroys fine-grained visual detail (numerical labels, OCR, small symbols) while preserving global layout and color, and keeps the visual token count unchanged so that the logprobs under $v$ and $\tilde{v}$ align token-for-token. The robust global image information that survives this degradation keeps $\teachert$ stable and on-distribution at inference time, so its difference from $\teacher$ reflects the missing fine-grained detail rather than an input-format mismatch.

\subsection{Motivating Observations}
\label{sec:motivation}

\begin{keyinsight}
Visual supervision in VLM rollouts is concentrated in a small minority of tokens, and these tokens carry the signal that teaches the student to attend to fine-grained visual detail. Masking their KL loss substantially degrades performance, while masking an equal fraction of low-VA or random tokens has negligible effect. Standard on-policy distillation weights all tokens uniformly and therefore dilutes this concentrated signal with surrounding language scaffolding.
\end{keyinsight}

We establish two empirical observations that motivate the token-level design of VA-OPD, with both measured directly on student-generated rollouts to reflect the student's own generation behavior.

\begin{figure}[t]
    \centering
    \begin{minipage}[b]{0.49\textwidth}
        \centering
        \includegraphics[width=\linewidth]{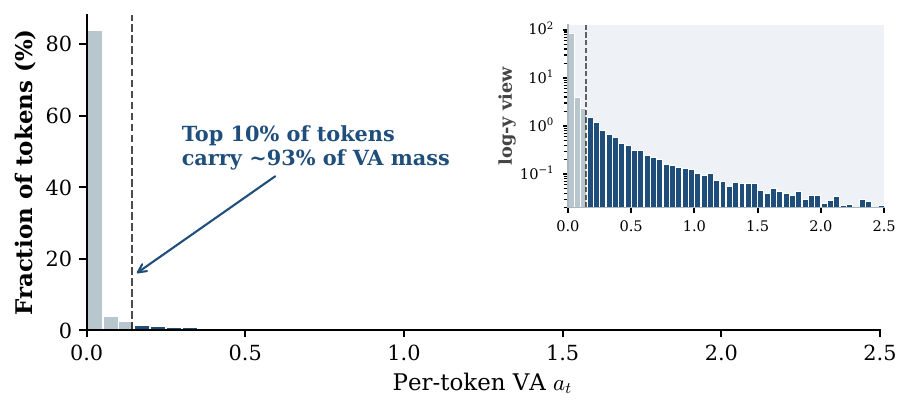}
    \end{minipage}
    \hfill
    \begin{minipage}[b]{0.49\textwidth}
        \centering
        \includegraphics[width=\linewidth]{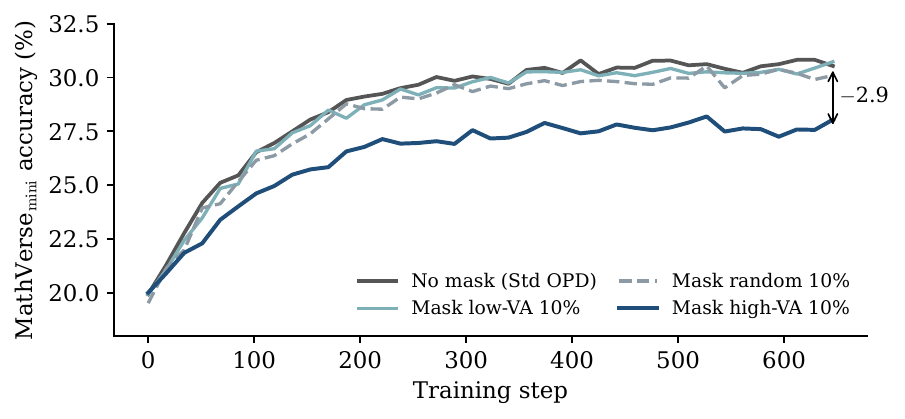}
    \end{minipage}
    \caption{\textbf{Motivating observations on student-generated rollouts.}
    \textbf{(a)}~VA is heavy-tailed: the top 10\% of tokens (right of the dashed line) carry ${\sim}93\%$ of total VA mass.
    \textbf{(b)}~MathVerse training curves for Standard OPD and three 10\%-token-masking variants; high-VA masking significantly drops accuracy, while low-VA or random masking is harmless.}
    \label{fig:motivation}
    \vspace{-3mm}
\end{figure}

\textbf{Observation 1: VA is sparse within a rollout.} For each student rollout, we compute $\va_t$ at every position and sort tokens by VA. \Cref{fig:motivation}(a) shows the resulting distribution, aggregated across $\sim$105k tokens of held-out rollouts from a Standard OPD student: the top 10\% of tokens carry $\sim$93\% of the total VA mass, while the rest have VA close to zero. This long-tail structure indicates that visual dependency in VLM rollouts is an inherently sparse phenomenon.

\textbf{Observation 2: High-VA tokens carry the visual supervision signal.} We design a controlled mask ablation: three Standard OPD variants (8B$\to$2B on Geometry3K) that differ only in which 10\% of tokens have their per-token KL \emph{masked out}: (i) a random 10\%, (ii) the lowest-VA 10\%, (iii) the highest-VA 10\%. All other training conditions are identical. \Cref{fig:motivation}(b) shows MathVerse$_{\rm mini}$ training curves. Masking low-VA tokens barely changes performance (within $\pm 0.1$); random masking causes only a $\sim$0.3-point drop; but masking the highest-VA 10\% drops MathVerse$_{\rm mini}$ by $\sim$2.9 points, despite this being only 10\% of the training signal. The high-VA minority therefore carries not just a disproportionate share of VA mass within each rollout, but also the supervision signal that most directly teaches the student to attend to fine-grained visual detail.

Together, these observations explain why standard on-policy distillation fails to strengthen the student's visual reliance: the visual supervision signal is concentrated in a small token minority, and under a uniform KL mean every token's gradient is scaled by $1/T$, so the concentrated signal is diluted by the surrounding scaffolding. We address this in \Cref{sec:method} by sorting tokens by VA and normalizing KL within high-VA and low-VA groups separately.

\section{Visual-Advantage On-Policy Distillation}
\label{sec:method}

\begin{figure}[t]
    \centering
    \includegraphics[width=\textwidth]{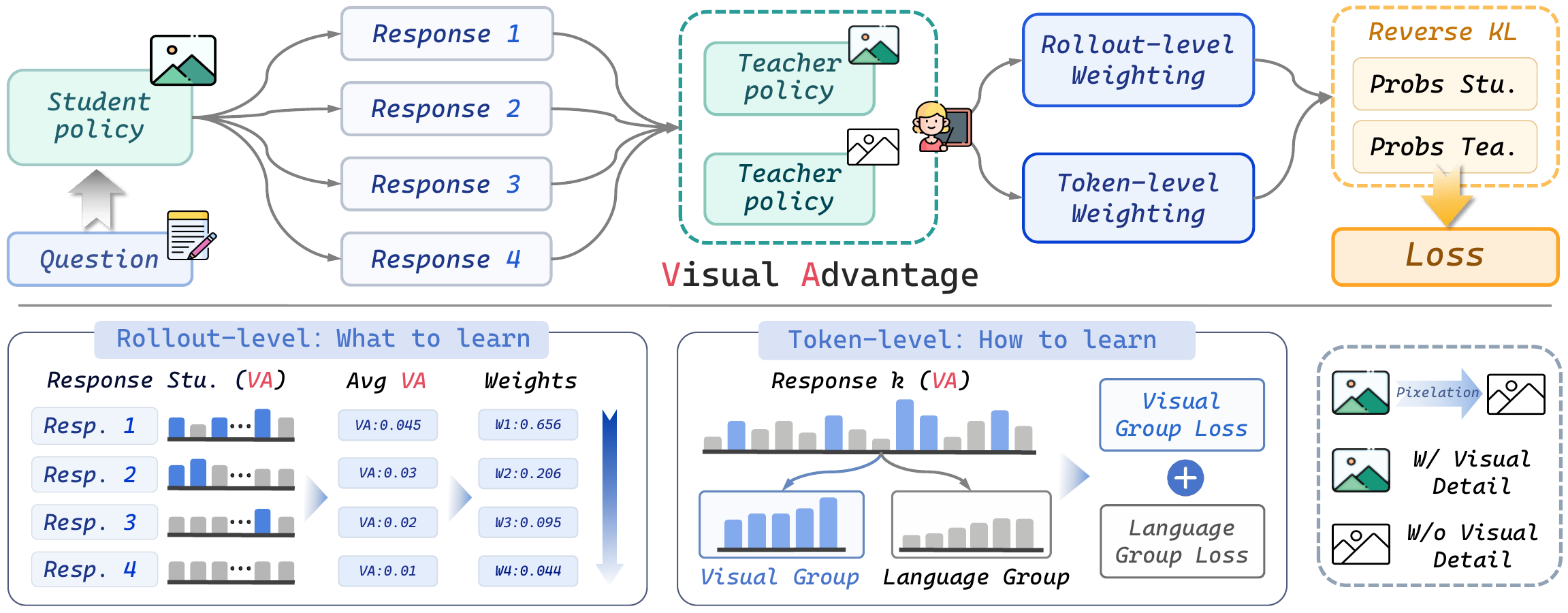}
    \caption{\textbf{Overview of Visual-Advantage On-Policy Distillation (VA-OPD).} The teacher scores each token with and without fine-grained visual detail; the log-prob difference gives the per-token visual advantage (VA). VA drives reverse-KL distillation at two granularities: rollout-level reweighting by relative VA (\emph{what to learn}) and token-level KL split into high- and low-VA groups (\emph{how to learn}).}
    \label{fig:pipeline}
    \vspace{-3mm}
\end{figure}

We propose \textbf{Visual-Advantage On-Policy Distillation (VA-OPD)} (\Cref{fig:pipeline}), which uses VA at two granularities. At the rollout level (\Cref{sec:c1}), each rollout is soft-weighted by its trajectory-averaged VA relative to its sibling rollouts under the same prompt, so that rollouts with stronger visual reliance contribute more to the gradient. At the token level (\Cref{sec:c2}), tokens within each rollout are sorted by VA into a high-VA group $V$ and a low-VA group $L$, and the per-token KL is averaged within each group separately before being summed. The full objective is given in \Cref{sec:objective}.

\subsection{Rollout-Level Reweighting}
\label{sec:c1}
Even when sampled from the same student under fixed $(v, q)$, sibling rollouts diverge in their token sequences and therefore in how much their tokens depend on fine-grained visual detail; their trajectory-averaged VA $\vabar^{(k)}$ varies non-trivially within a $K$-rollout group. We exploit this within-group spread by assigning each rollout a soft weight derived from its visual advantage relative to its $K-1$ siblings, so that more gradient mass flows to rollouts with stronger visual reliance.

For each rollout $y^{(k)}$, we compute the trajectory-averaged VA and normalize it within the $K$ sibling rollouts:
\begin{equation}
    \vabar^{(k)} = \frac{1}{T^{(k)}} \sum_{t=1}^{T^{(k)}} \va_t^{(k)},
    \qquad
    \hat{z}^{(k)} = \frac{\vabar^{(k)} - \mu}{\sigma + \epsilon},
    \label{eq:relva}
\end{equation}
where $\mu$ and $\sigma$ are the mean and standard deviation of $\{\vabar^{(j)}\}_{j=1}^{K}$, and $\epsilon$ is a small constant for numerical stability. A positive $\hat{z}^{(k)}$ indicates that the rollout is more vision-engaged than its siblings within the same candidate set; a negative value indicates it is relatively less so. We then convert $\hat{z}^{(k)}$ into rollout-level distillation weights via a temperature-controlled softmax with $\tau = 1.0$:
\begin{equation}
    w^{(k)} = \frac{\exp\!\left(\hat{z}^{(k)} / \tau\right)}{\sum_{j=1}^{K} \exp\!\left(\hat{z}^{(j)} / \tau\right)},
    \qquad \sum_{k=1}^{K} w^{(k)} = 1.
    \label{eq:weights}
\end{equation}

\subsection{Token-Level Grouped KL}
\label{sec:c2}
Within each rollout, vision-critical tokens form a small minority; most tokens are language scaffolding. Under a uniform KL mean over all $T$ tokens, every token's gradient is scaled by $1/T$, so the magnitude of the visual supervision shrinks with the length of the surrounding scaffolding. We address this by sorting each rollout's tokens by VA, treating the top fraction as a high-VA group and the rest as a low-VA group, and averaging per-token KL within each group separately.

For each rollout $y^{(k)}$ of length $T^{(k)}$, we rank its tokens by VA $\{\va_t^{(k)}\}_{t=1}^{T^{(k)}}$ and place the top-$p_v$ fraction in the high-VA group $V^{(k)}$; the remaining tokens form the low-VA group $L^{(k)}$. The hyperparameter $p_v \in (0, 1)$ controls the relative size of $V$ and we fix $p_v = 0.2$ for all experiments. The per-rollout loss is a weighted sum of the two within-group KL averages:
\begin{equation}
    \cL^{(k)}_{\text{group}} = \lambda \cdot \frac{1}{|V^{(k)}|} \sum_{t \in V^{(k)}} \KL_t \;+\; (1-\lambda) \cdot \frac{1}{|L^{(k)}|} \sum_{t \in L^{(k)}} \KL_t,
    \label{eq:vlkl}
\end{equation}
where $\KL_t = \KL\!\left(\student(\cdot \mid v, q, y_{<t}) \;\|\; \teacher(\cdot \mid v, q, y_{<t})\right)$ is the per-token reverse KL from \Cref{eq:standard_kl}, and $\lambda \in (0,1)$ controls the high-VA group's weight. We use $\lambda = 0.5$ as the default. Since each group is size-normalized, the high-VA group's gradient contribution depends only on $|V|$ and $\lambda$, not on $|L|$ or total rollout length. The degraded image $\tilde{v}$ is used only for VA computation; the KL target in \Cref{eq:vlkl} always uses the teacher distribution conditioned on the original image $v$.

\subsection{Overall Objective}
\label{sec:objective}
Combining rollout-level weights with token-level grouped KL gives the per-prompt VA-OPD loss:
\begin{equation}
    \Lpgd(x) = \sum_{k=1}^{K} w^{(k)} \left[ \lambda \cdot \frac{1}{|V^{(k)}|} \sum_{t \in V^{(k)}} \KL_t + (1-\lambda) \cdot \frac{1}{|L^{(k)}|} \sum_{t \in L^{(k)}} \KL_t \right],
    \label{eq:total}
\end{equation}
with full training objective $\Lpgd = \mathbb{E}_{x \sim \mathcal{D}} \left[ \Lpgd(x) \right]$. VA-OPD needs no additional annotations, reward models, or student rollout changes; the teacher performs only one extra forward pass per rollout without fine-grained visual detail to supply VA signals, leaving inference unaffected.

\section{Experiments}
\label{sec:experiments}

We ask: (Q1)~does VA-OPD beat Standard OPD and RL baselines on accuracy (\Cref{sec:main_results}); (Q2)~does it generalize across scales and data (\Cref{sec:scaling}); (Q3)~which components drive the gain (\Cref{sec:ablation}); and (Q4)~is VA-OPD's added per-step compute justified by its accuracy gain (\Cref{sec:efficiency})?

\subsection{Experimental Setup}
\label{sec:setup}

\noindent{\textbf{Models and data.}}
Our primary setting uses Qwen3-VL-8B-Instruct~\citep{bai2025qwen3} as the teacher and Qwen3-VL-2B-Instruct as the student, trained on Geometry3K~\citep{lu2021inter} (${\sim}2.1$K problems). We further test generalization across teacher sizes (Qwen3-VL-4B$\to$2B and Qwen3-VL-32B$\to$2B) and on a larger dataset ViRL39K~\citep{wang2025vl} (${\sim}39$K problems). The teacher distribution $\teachert$ used to compute VA follows the construction in \Cref{sec:va}.

\noindent{\textbf{Training.}}
We train for 5 epochs using AdamW with $K=4$ rollouts per prompt and batch size 16. VA-OPD defaults are $\tau = 1.0$ (rollout-reweighting softmax temperature), $\lambda = 0.5$ (high-VA group weight), $p_v = 0.2$ (high-VA fraction), and pixelation ratio $0.10$; these are fixed across all configurations with no per-configuration tuning, and all methods within a configuration share identical initialization, data, optimizer, and evaluation protocol, making comparisons attributable to the training objective itself. Full training, hyperparameter, and compute-fairness details are in \Cref{app:impl}.

\noindent{\textbf{Baselines and benchmarks.}}
We compare VA-OPD against six baselines reimplemented under identical conditions: Base (the un-distilled student evaluated directly), CoT-SFT, Off-policy KD~\citep{hinton2015distilling}, Standard OPD~\citep{gu2024minillm} (\Cref{eq:standard_kl}), GRPO~\citep{shao2024deepseekmath}, and PAPO~\citep{wang2025perception}. We evaluate on eight benchmarks grouped into mathematical reasoning (WeMath~\citep{qiao2407we}, MathVista~\citep{lu2023mathvista}, MathVerse~\citep{zhang2024mathverse}) and visual understanding (HallusionBench~\citep{guan2024hallusionbench}, AI2D~\citep{kembhavi2016diagram}, MMMU~\citep{yue2024mmmu}, MMStar~\citep{chen2024we}, OCRBench~\citep{liu2024ocrbench}). We report avg@8 at temperature 1.0 under official scoring protocols (GPT-4o judge where applicable).

\subsection{Main Results}
\label{sec:main_results}

\begin{table}[t]
\centering
\vspace{-6pt}
\caption{\textbf{Main results on eight benchmarks.}
Primary setting: Qwen3-VL-8B$\to$Qwen3-VL-2B, trained on Geometry3K. Each method is reported with its best checkpoint. Scores are avg@8 (temperature 1.0); per-benchmark metric conventions follow each benchmark's official protocol (\Cref{app:eval_protocol}). The highest score in each column is shown in bold.}
\label{tab:main}
\vspace{4pt}
\resizebox{\textwidth}{!}{
\begin{tabular}{l ccc c ccccc c}
\toprule
\multirow{2}{*}{\textbf{Method}} & \multicolumn{4}{c}{\cellcolor{blue!15}\textbf{Math Reasoning}} & \multicolumn{6}{c}{\cellcolor{orange!18}\textbf{Visual Understanding}} \\
\cmidrule(lr){2-5} \cmidrule(lr){6-11}
& WeMath & MathVista & MathVerse & \cellcolor{gray!12}\textbf{Avg} & HalluB & AI2D & MMMU & MMStar & OCRBench & \cellcolor{gray!12}\textbf{Avg} \\
\midrule
Base & 36.8 & 63.9 & 19.6 & \cellcolor{gray!12}40.1 & 52.6 & 77.8 & 49.1 & 56.1 & 85.9 & \cellcolor{gray!12}64.3 \\
CoT-SFT & 39.7 & 64.4 & 23.9 & \cellcolor{gray!12}42.7 & 51.0 & 75.7 & 48.7 & 57.1 & 86.2 & \cellcolor{gray!12}63.7 \\
Off-policy KD & 38.9 & 65.3 & 24.3 & \cellcolor{gray!12}42.8 & 52.3 & 76.0 & 49.3 & 57.4 & 85.5 & \cellcolor{gray!12}64.1 \\
GRPO & 44.1 & 64.6 & 24.9 & \cellcolor{gray!12}44.5 & 54.0 & 77.5 & \textbf{52.9} & 57.8 & 84.8 & \cellcolor{gray!12}65.4 \\
PAPO & 44.8 & 64.5 & 25.8 & \cellcolor{gray!12}45.0 & 54.0 & 77.8 & 52.7 & 58.0 & 84.7 & \cellcolor{gray!12}65.4 \\
OPD & 43.3 & 63.7 & 29.1 & \cellcolor{gray!12}45.4 & 52.0 & 75.8 & 50.9 & 59.7 & 84.7 & \cellcolor{gray!12}64.6 \\
\midrule
\textbf{VA-OPD (ours)} & \textbf{46.6} & \textbf{66.4} & \textbf{31.9} & \cellcolor{green!20}\textbf{48.3}\,\tup{2.9} & \textbf{54.5} & \textbf{78.2} & 51.5 & \textbf{59.9} & \textbf{86.4} & \cellcolor{green!20}\textbf{66.1}\,\tup{1.5} \\
\bottomrule
\end{tabular}
}
\vspace{-4pt}
\end{table}

\begin{table}[t]
\centering
\vspace{-6pt}
\caption{\textbf{Standard OPD vs.\ VA-OPD across four configurations} (avg@8). All configurations use the Qwen3-VL family; per-benchmark metric conventions follow \Cref{app:eval_protocol}.}
\label{tab:scaling_summary}
\vspace{4pt}
\small
\setlength{\tabcolsep}{3.5pt}
\resizebox{\textwidth}{!}{
\begin{tabular}{ll l ccc c ccccc c}
\toprule
\multirow{2}{*}{\textbf{T$\to$S}} & \multirow{2}{*}{\textbf{Data}} & \multirow{2}{*}{\textbf{Method}} & \multicolumn{4}{c}{\cellcolor{blue!15}\textbf{Math Reasoning}} & \multicolumn{6}{c}{\cellcolor{orange!18}\textbf{Visual Understanding}} \\
\cmidrule(lr){4-7} \cmidrule(lr){8-13}
& & & WeMath & MathVista & MathVerse & \cellcolor{gray!12}\textbf{Avg} & HalluB & AI2D & MMMU & MMStar & OCRBench & \cellcolor{gray!12}\textbf{Avg} \\
\midrule
\multirow{2}{*}{4B$\to$2B} & \multirow{2}{*}{Geo3K}
  & OPD & 43.5 & 65.4 & 26.9 & \cellcolor{gray!12}45.3 & 52.5 & 75.0 & 49.2 & 60.4 & 84.8 & \cellcolor{gray!12}64.4 \\
& & \textbf{VA-OPD} & \textbf{45.9} & \textbf{67.3} & \textbf{29.0} & \cellcolor{green!20}\textbf{47.4}\,\tup{2.1} & \textbf{53.7} & \textbf{75.8} & \textbf{49.9} & \textbf{61.1} & \textbf{85.5} & \cellcolor{green!20}\textbf{65.2}\,\tup{0.8} \\
\midrule
\multirow{2}{*}{8B$\to$2B} & \multirow{2}{*}{Geo3K}
  & OPD & 43.3 & 63.7 & 29.1 & \cellcolor{gray!12}45.4 & 52.0 & 75.8 & 50.9 & 59.7 & 84.7 & \cellcolor{gray!12}64.6 \\
& & \textbf{VA-OPD} & \textbf{46.6} & \textbf{66.4} & \textbf{31.9} & \cellcolor{green!20}\textbf{48.3}\,\tup{2.9} & \textbf{54.5} & \textbf{78.2} & \textbf{51.5} & \textbf{59.9} & \textbf{86.4} & \cellcolor{green!20}\textbf{66.1}\,\tup{1.5} \\
\midrule
\multirow{2}{*}{32B$\to$2B} & \multirow{2}{*}{Geo3K}
  & OPD & 54.1 & 67.6 & 31.5 & \cellcolor{gray!12}51.1 & 52.3 & 77.0 & 50.0 & 62.5 & 84.7 & \cellcolor{gray!12}65.3 \\
& & \textbf{VA-OPD} & \textbf{57.5} & \textbf{70.8} & \textbf{36.1} & \cellcolor{green!20}\textbf{54.8}\,\tup{3.7} & \textbf{55.1} & \textbf{79.5} & \textbf{51.0} & \textbf{64.0} & \textbf{86.9} & \cellcolor{green!20}\textbf{67.3}\,\tup{2.0} \\
\midrule
\multirow{2}{*}{8B$\to$2B} & \multirow{2}{*}{ViRL39K}
  & OPD & 46.1 & 63.3 & 29.7 & \cellcolor{gray!12}46.4 & 53.6 & 75.4 & 51.4 & 61.8 & 85.2 & \cellcolor{gray!12}65.5 \\
& & \textbf{VA-OPD} & \textbf{50.3} & \textbf{67.5} & \textbf{32.7} & \cellcolor{green!20}\textbf{50.2}\,\tup{3.8} & \textbf{55.8} & \textbf{79.5} & \textbf{53.2} & \textbf{64.0} & \textbf{87.5} & \cellcolor{green!20}\textbf{68.0}\,\tup{2.5} \\
\bottomrule
\end{tabular}
}
\vspace{-4pt}
\end{table}

\Cref{tab:main} reports the primary comparison of all seven methods across eight benchmarks (8B$\to$2B, trained on Geometry3K only), with model scale and training data held fixed.

\noindent{\textbf{VA-OPD consistently outperforms Standard OPD across all benchmarks.}}
Compared to Standard OPD, the strongest reproduced distillation baseline, VA-OPD improves on all eight benchmarks, with no benchmark dropping relative to Standard OPD, spanning both mathematical reasoning and visual-understanding tasks: WeMath $+3.3$, MathVista $+2.7$, MathVerse $+2.8$, HallusionBench $+2.5$, AI2D $+2.4$, MMMU $+0.6$, MMStar $+0.2$, and OCRBench $+1.7$. The aggregate effect is $+2.9$ on Math Avg and $+1.5$ on Visual Avg, showing that the advantage is broad rather than driven by one outlier; within visual understanding, gains concentrate on HallusionBench and AI2D, where answers hinge on careful image reading, suggesting that encouraging stronger visual dependency during distillation reduces visual hallucination while improving diagram understanding.

\noindent{\textbf{The headline gain pattern tracks visual dependency.}}
The benchmarks on which VA-OPD improves the most are precisely those that most directly depend on fine-grained visual detail, namely HallusionBench (image-grounded hallucination probing) and AI2D (diagram-based reasoning), whereas benchmarks that admit text-only shortcuts (e.g., MMMU, MMStar) see noticeably smaller gains. This pattern, in turn, is consistent with VA-OPD's mechanism: by sorting tokens by VA, the method concentrates the distillation signal on positions whose teacher predictions actually require fine-grained visual detail, while leaving language-driven tokens essentially unchanged.

\subsection{Generalization across Scales and Data}
\label{sec:scaling}

\Cref{tab:scaling_summary} extends the comparison to four configurations that vary training-data scale and teacher size (with the student fixed at 2B).

\noindent{\textbf{Larger training corpus does not dilute the gain.}}
Holding the 8B$\to$2B teacher--student pair fixed and switching from Geometry3K (${\sim}2.1$K problems) to the substantially larger ViRL39K (${\sim}39$K problems), which isolates data scale rather than model capacity in this comparison, VA-OPD's Math Avg gain over Standard OPD grows from $+2.9$ to $+3.8$ while Visual Avg grows from $+1.5$ to $+2.5$. A roughly $20\times$ increase in training data therefore, under this fixed model scale, does not dilute the visual-dependency signal or average it away, but amplifies it, consistent with broader visual coverage producing more high-VA tokens for VA-OPD to exploit.

\noindent{\textbf{Gain holds across teacher sizes.}}
Holding the student fixed at 2B and varying the teacher across 4B, 8B, and 32B on Geometry3K, thereby isolating teacher capacity as the varying factor, the gain of VA-OPD over Standard OPD stays clearly positive at all three scales: $+$2.1/$+$0.8 at 4B, $+$2.9/$+$1.5 at 8B, and $+$3.7/$+$2.0 at 32B for Math/Visual Avg, with stronger teachers producing richer counterfactual signals for VA-OPD to extract, showing that the effect is not scale-specific.

\subsection{Ablation Study and Controls}
\label{sec:ablation}

\begin{figure}[t]
    \centering
    \vspace{-6pt}
    \includegraphics[width=\textwidth]{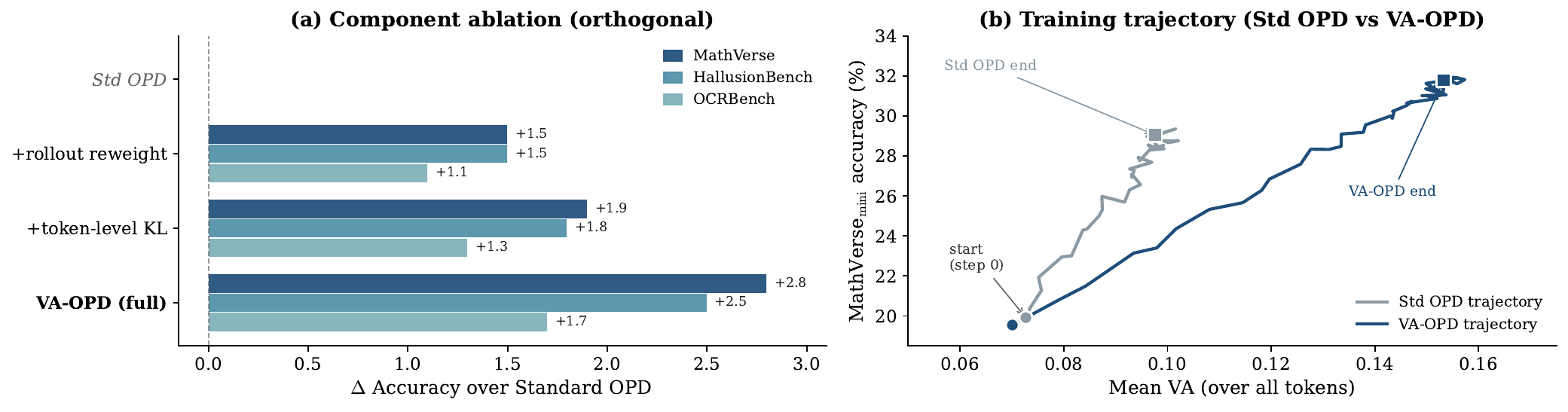}
    \caption{\textbf{Ablation and training-trajectory analysis.}
    (a)~Per-benchmark $\Delta$ over Standard OPD for two VA-OPD components alone and combined, separating their individual and joint contributions: rollout-level reweighting only, token-level grouped KL only, and full VA-OPD.
    (b)~Training trajectories in $(\text{mean VA over all tokens}, \text{MathVerse},\text{accuracy})$ space throughout training for Standard OPD vs.\ VA-OPD students. Circles mark the start (initial Qwen3-VL-2B-Instruct), squares the end of training. The trajectories move in distinctly different directions: Standard OPD climbs nearly vertically (accuracy rises while mean VA stays near the start), whereas VA-OPD moves diagonally up-right (accuracy and mean VA rise together, indicating coupled improvement).}
    \label{fig:ablation}
    \vspace{-6pt}
\end{figure}

We validate VA-OPD along two complementary axes (\Cref{fig:ablation}): (a)~a component ablation on MathVerse (vision-intensive math), HallusionBench (hallucination), and OCRBench (OCR) that isolates each VA-OPD component and its contribution to overall gains; (b)~a training trajectory comparison that plots each student's path through $(\text{VA}, \text{accuracy})$ space across training, contrasting Standard OPD with VA-OPD and revealing whether accuracy and VA co-evolve.

% \noindent{\textbf{Both components are necessary.}}
% Removing either VA-OPD component (rollout reweighting or token-level grouped KL) degrades performance, so neither mechanism is sufficient alone; only their combination delivers the full VA-OPD gain. Each component alone nevertheless recovers a sizable fraction of the gain, indicating that both contribute nontrivially. Full VA-OPD reaches $+$2.8, $+$2.5, and $+$1.7 on MathVerse, HallusionBench, and OCRBench respectively, peaking on MathVerse where solving the problem requires reading off precise diagram quantities from the image.

\noindent{\textbf{Both components are necessary.}}
Removing either VA-OPD component (rollout reweighting or token-level grouped KL) measurably degrades performance, confirming that neither mechanism is sufficient on its own; only their joint combination delivers the full VA-OPD gain. Each component alone nevertheless recovers a sizable fraction of the improvement, indicating that the two contribute through nontrivial and largely complementary pathways. Full VA-OPD reaches $+$2.8, $+$2.5, and $+$1.7 on MathVerse, HallusionBench, and OCRBench respectively, peaking on MathVerse where solving the problem demands reading off precise diagram quantities directly from the image, exactly the regime where visual grounding matters most.

\noindent{\textbf{VA-OPD raises student visual reliance during training.}}
\Cref{fig:ablation}(b) plots each student's path in $(\text{mean VA}, \text{MathVerse},\text{accuracy})$ space, where mean VA is averaged over \emph{all} tokens rather than selected high-VA positions. Standard OPD traces a nearly \emph{vertical} path: accuracy rises from $19.6$ to $29.1$ while the all-token mean VA only inches up ($0.07 \to 0.10$). VA-OPD instead traces a \emph{diagonal} path: accuracy climbs higher (to $31.9$) \emph{and} the all-token mean VA grows substantially ($0.07 \to 0.16$). The two trajectory shapes make this mechanism visible at a glance under this stricter all-token diagnostic: Standard OPD buys accuracy without stronger visual dependency, whereas VA-OPD's accuracy gain is coupled with a genuine increase in visual reliance across the rollout.

\subsection{Training Efficiency}
\label{sec:efficiency}

\begin{wrapfigure}{r}{0.40\textwidth}
\centering
\vspace{-16pt}
\setlength{\abovecaptionskip}{3pt}
\setlength{\belowcaptionskip}{0pt}
\includegraphics[width=\linewidth]{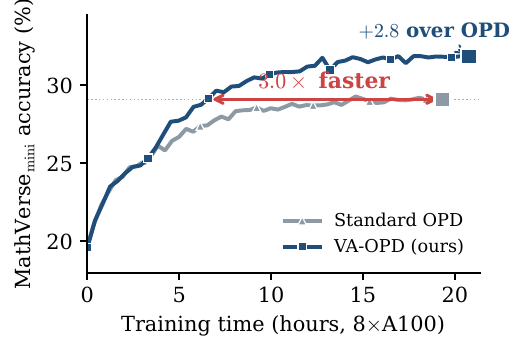}
\caption{\textbf{MathVerse accuracy vs.\ training time on 8$\times$A100 (Qwen3-VL-8B$\to$2B, Geo3K).} The horizontal arrow marks how much faster VA-OPD reaches Standard OPD's final accuracy.}
\label{fig:efficiency}
\vspace{-12pt}
\end{wrapfigure}

% \Cref{fig:efficiency} plots MathVerse accuracy against wall-clock time on the primary 8B$\to$2B Geo3K configuration. VA-OPD's curve sits above Standard OPD's at every time point, so the gain shows up in two complementary ways: (i)~at a fixed accuracy target, VA-OPD matches Standard OPD's final accuracy of $29.1$ in only ${\sim}6.5$~h versus $19.3$~h, a ${\sim}3.0\times$ wall-clock speedup; (ii)~at a fixed time budget, VA-OPD plateaus at a higher ceiling of $31.9$, $+2.8$ above Standard OPD's endpoint. The two curves overlap for the first ${\sim}3$~h and diverge only afterward, suggesting that the visual-dependency signal becomes effective once the student first matches the teacher's surface distribution; from then on, high-VA tokens are the bottleneck for further gains, exactly where VA-OPD redirects the gradient.

\Cref{fig:efficiency} plots MathVerse accuracy against wall-clock time on the primary 8B$\to$2B Geo3K configuration. VA-OPD's curve never falls below Standard OPD's; the two coincide for roughly the first ${\sim}3$~h and diverge only afterward, so the resulting gain manifests in two complementary ways: (i)~at a fixed accuracy target, VA-OPD matches Standard OPD's final accuracy of $29.1$ in only ${\sim}6.5$~h versus $19.3$~h, yielding a ${\sim}3.0\times$ wall-clock speedup; (ii)~at a fixed time budget, VA-OPD plateaus at a higher ceiling of $31.9$, fully $+2.8$ points above Standard OPD's endpoint. The early-stage coincidence suggests that the visual-dependency signal becomes effective once the student has first matched the teacher's surface distribution; from then on, high-VA tokens emerge as the dominant bottleneck for further gains, which is precisely where VA-OPD redirects the gradient to sustain continued improvement.
\section{Qualitative Token-Level VA Visualization}
\label{sec:analysis}

\begin{figure}[t]
    \centering
    \includegraphics[width=\textwidth]{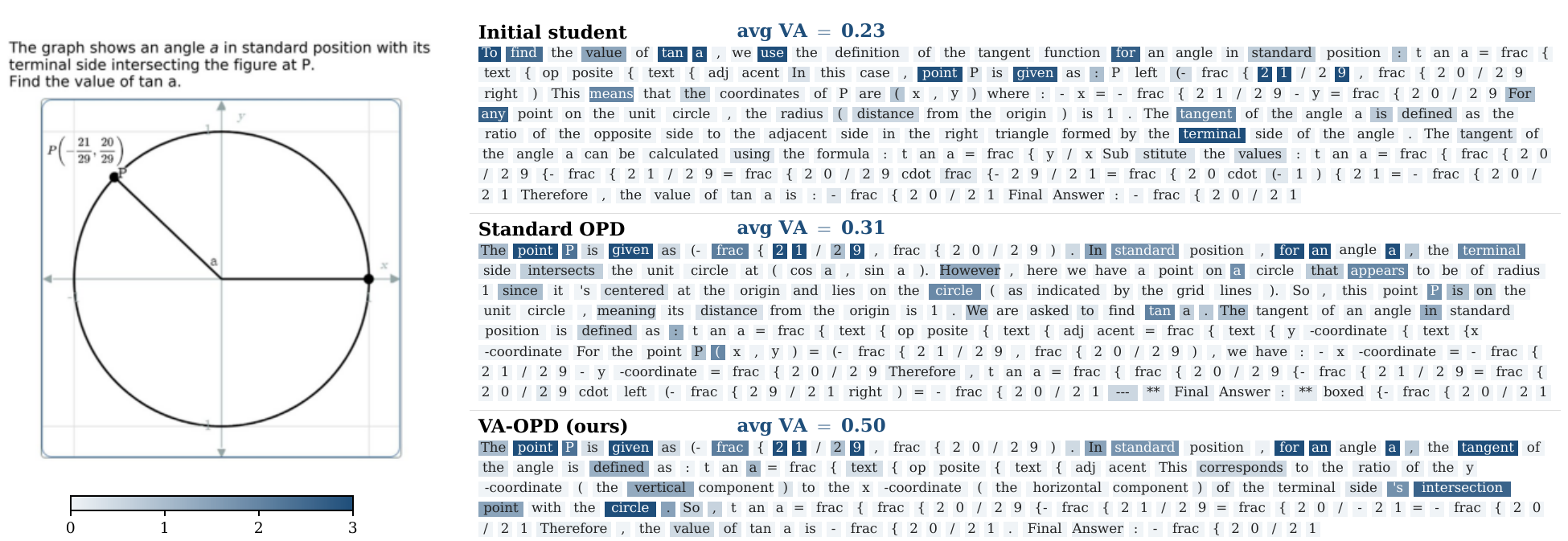}
    \caption{\textbf{Token-level VA visualization} on a representative geometry rollout. Each token is colored by its VA under three students at different stages: the initial student (Base, before distillation), the Standard OPD student, and the VA-OPD student (darker $=$ higher $\mathrm{VA}_t$). The visually critical tokens, namely the numerical values read from the diagram, start pale under the initial student, remain pale under Standard OPD, and become visually dependent only under VA-OPD.}
    \label{fig:va_heatmap}
\end{figure}

\Cref{fig:va_heatmap} inspects the visual advantage at the token level on a held-out geometry rollout. The initial student (before any distillation) exhibits uniformly low VA on the vision-critical tokens; Standard OPD does not alter this pattern, and those tokens stay pale even though the output of the student now matches that of the teacher; only VA-OPD develops clear VA hotspots on the vision-critical tokens, which visually confirms that VA-OPD trains the student to \emph{read the image} rather than mimic surface text tokens. Concretely, hotspots concentrate on numerical values that must be extracted from the diagram, including specific angle measures and side lengths, while language-template tokens such as connectors, fillers, and equation rewrites stay pale across all three students. This per-token contrast complements the controlled mask experiment in \Cref{sec:motivation}: the mask experiment showed that a small VA-aware token subset carries the entire visual supervision signal, and the heatmap shows this subset is precisely the vision-critical tokens dependent on fine-grained visual detail. Together, the quantitative ablation and qualitative heatmap converge on the same conclusion: VA-OPD redirects gradient mass toward tokens that genuinely depend on fine-grained visual detail.

\section{Related Work}
\label{sec:related}

\paragraph{Visual dependency in VLMs.}
VLMs often fall back on language priors instead of attending to fine-grained visual detail~\citep{favero2024multi,leng2024mitigating,ghosh2024visual,xie2024v,li2025self}. Mitigations include inference-time contrastive decoding~\citep{leng2024mitigating,favero2024multi,wang2024mitigating,ghosh2024visual}, preference-based fine-tuning~\citep{xie2024v,gu2024token}, and RL-based methods that inject perception into the policy gradient~\citep{wang2025perception,huang2025spotlight,li2026rethinking,liu2025noisyrollout}.  Based on this, we propose VA-OPD, which applies VA at two complementary granularities: rollout-level reweighting by trajectory-averaged VA, and token-level KL averaged separately within high-VA and low-VA groups.

\paragraph{On-policy distillation for LLMs.}
GKD~\citep{agarwal2024policy} and MiniLLM~\citep{gu2024minillm} established on-policy distillation via teacher matching on student rollouts. Subsequent work refines the divergence~\citep{ko2024distillm,wu2025rethinking,jin2026entropy,yang2026learning}, the sampling strategy~\citep{xu2024speculative}, or identifies important tokens by teacher penalty or entropy~\citep{lu2025policy,wang2025beyond}. These methods all operate in the text-only setting with token-uniform supervision; by contrast, VA-OPD extends this paradigm to VLMs and shows that such uniform treatment misses the sparse concentration of visual supervision within visually grounded VLM rollouts.

\paragraph{Positioning.}
VA-OPD sits at the intersection of these two lines, and to our knowledge is the first to embed per-token reliance on fine-grained visual detail directly into the on-policy distillation target for VLMs. Unlike inference-time methods, it modifies the training target with no test-time overhead; unlike RL-based perception methods, it leverages teacher supervision rather than verifiable rewards.
\section{Conclusion}
\label{sec:conclusion}

We showed that standard on-policy distillation for VLMs fails to strengthen the student's reliance on fine-grained visual detail: the visual supervision concentrated on a small minority of tokens is diluted by uniform-weighted KL averaging across the surrounding language scaffolding. To observe this insensitivity, we introduced visual advantage (VA), a teacher-relative counterfactual measuring token-level dependency on fine-grained visual detail. Building on VA, we proposed Visual-Advantage On-Policy Distillation (VA-OPD), which uses VA at two granularities: rollout-level reweighting that redistributes gradient mass toward rollouts most reliant on fine-grained visual detail, and token-level KL averaged separately within high-VA and low-VA groups so the concentrated visual signal is not diluted by the surrounding scaffolding. Across three teacher sizes (4B, 8B, and 32B), two training corpora (Geometry3K and ViRL39K), and eight benchmarks, VA-OPD consistently improves over standard on-policy distillation, with the gain amplifying as the teacher and the training corpus scale. The teacher-scored VA on the student's rollouts rises in lockstep with accuracy during VA-OPD training, indicating gains come with genuine visual reliance rather than output mimicry alone.

\noindent{\textbf{Limitations.}}
VA measures teacher-relative counterfactual sensitivity, which is a useful proxy for visual dependency but not a direct measurement of internal perceptual processes such as attention allocation or feature grounding. The strongest gains appear on vision-intensive tasks; improvements on benchmarks where many questions are solvable from text alone (e.g., MMStar) are more modest.

\noindent{\textbf{Future work.}}
VA-OPD could more directly complement RL-based visual perception methods: it strengthens visual dependency during distillation, while RL optimizes task reward. Extending VA-OPD to region- or object-level visual dependency is also promising for future work.

\bibliography{references}

@article{hinton2015distilling,
  title={Distilling the knowledge in a neural network},
  author={Hinton, Geoffrey and Vinyals, Oriol and Dean, Jeff},
  journal={arXiv preprint arXiv:1503.02531},
  year={2015}
}

@inproceedings{agarwal2024policy,
  title={On-policy distillation of language models: Learning from self-generated mistakes},
  author={Agarwal, Rishabh and Vieillard, Nino and Zhou, Yongchao and Stanczyk, Piotr and Garea, Sabela Ramos and Geist, Matthieu and Bachem, Olivier},
  booktitle={The twelfth international conference on learning representations},
  year={2024}
}

@inproceedings{gu2024minillm,
  title={Minillm: Knowledge distillation of large language models},
  author={Gu, Yuxian and Dong, Li and Wei, Furu and Huang, Minlie},
  booktitle={The twelfth international conference on learning representations},
  year={2024}
}

@article{ko2024distillm,
  title={Distillm: Towards streamlined distillation for large language models},
  author={Ko, Jongwoo and Kim, Sungnyun and Chen, Tianyi and Yun, Se-Young},
  journal={arXiv preprint arXiv:2402.03898},
  year={2024}
}

@inproceedings{wu2025rethinking,
  title={Rethinking kullback-leibler divergence in knowledge distillation for large language models},
  author={Wu, Taiqiang and Tao, Chaofan and Wang, Jiahao and Yang, Runming and Zhao, Zhe and Wong, Ngai},
  booktitle={Proceedings of the 31st International Conference on Computational Linguistics},
  pages={5737--5755},
  year={2025}
}

@article{xu2024speculative,
  title={Speculative knowledge distillation: Bridging the teacher-student gap through interleaved sampling},
  author={Xu, Wenda and Han, Rujun and Wang, Zifeng and Le, Long T and Madeka, Dhruv and Li, Lei and Wang, William Yang and Agarwal, Rishabh and Lee, Chen-Yu and Pfister, Tomas},
  journal={arXiv preprint arXiv:2410.11325},
  year={2024}
}

@article{jin2026entropy,
  title={Entropy-Aware On-Policy Distillation of Language Models},
  author={Jin, Woogyeol and Min, Taywon and Yang, Yongjin and Kadhe, Swanand Ravindra and Zhou, Yi and Wei, Dennis and Baracaldo, Nathalie and Lee, Kimin},
  journal={arXiv preprint arXiv:2603.07079},
  year={2026}
}

@article{yang2026learning,
  title={Learning beyond teacher: Generalized on-policy distillation with reward extrapolation},
  author={Yang, Wenkai and Liu, Weijie and Xie, Ruobing and Yang, Kai and Yang, Saiyong and Lin, Yankai},
  journal={arXiv preprint arXiv:2602.12125},
  year={2026}
}

@misc{lu2025policy,
  title={On-Policy Distillation},
  author={Lu, Kevin},
  howpublished={Thinking Machines Lab Blog (Connectionism)},
  url={https://thinkingmachines.ai/blog/on-policy-distillation/},
  year={2025},
  note={Published 2025-10-27}
}

@article{wang2025beyond,
  title={Beyond the 80/20 rule: High-entropy minority tokens drive effective reinforcement learning for llm reasoning},
  author={Wang, Shenzhi and Yu, Le and Gao, Chang and Zheng, Chujie and Liu, Shixuan and Lu, Rui and Dang, Kai and Chen, Xionghui and Yang, Jianxin and Zhang, Zhenru and others},
  journal={arXiv preprint arXiv:2506.01939},
  year={2025}
}

@article{shao2024deepseekmath,
  title={Deepseekmath: Pushing the limits of mathematical reasoning in open language models},
  author={Shao, Zhihong and Wang, Peiyi and Zhu, Qihao and Xu, Runxin and Song, Junxiao and Bi, Xiao and Zhang, Haowei and Zhang, Mingchuan and Li, YK and Wu, Yang and others},
  journal={arXiv preprint arXiv:2402.03300},
  year={2024}
}

@inproceedings{leng2024mitigating,
  title={Mitigating object hallucinations in large vision-language models through visual contrastive decoding},
  author={Leng, Sicong and Zhang, Hang and Chen, Guanzheng and Li, Xin and Lu, Shijian and Miao, Chunyan and Bing, Lidong},
  booktitle={Proceedings of the IEEE/CVF Conference on Computer Vision and Pattern Recognition},
  pages={13872--13882},
  year={2024}
}

@inproceedings{favero2024multi,
  title={Multi-modal hallucination control by visual information grounding},
  author={Favero, Alessandro and Zancato, Luca and Trager, Matthew and Choudhary, Siddharth and Perera, Pramuditha and Achille, Alessandro and Swaminathan, Ashwin and Soatto, Stefano},
  booktitle={Proceedings of the IEEE/CVF Conference on Computer Vision and Pattern Recognition},
  pages={14303--14312},
  year={2024}
}

@inproceedings{wang2024mitigating,
  title={Mitigating hallucinations in large vision-language models with instruction contrastive decoding},
  author={Wang, Xintong and Pan, Jingheng and Ding, Liang and Biemann, Chris},
  booktitle={Findings of the Association for Computational Linguistics: ACL 2024},
  pages={15840--15853},
  year={2024}
}

@article{ghosh2024visual,
  title={Visual description grounding reduces hallucinations and boosts reasoning in lvlms},
  author={Ghosh, Sreyan and Evuru, Chandra Kiran Reddy and Kumar, Sonal and Tyagi, Utkarsh and Nieto, Oriol and Jin, Zeyu and Manocha, Dinesh},
  journal={arXiv preprint arXiv:2405.15683},
  year={2024}
}

@inproceedings{xie2024v,
  title={V-dpo: Mitigating hallucination in large vision language models via vision-guided direct preference optimization},
  author={Xie, Yuxi and Li, Guanzhen and Xu, Xiao and Kan, Min-Yen},
  booktitle={Findings of the Association for Computational Linguistics: EMNLP 2024},
  pages={13258--13273},
  year={2024}
}

@article{gu2024token,
  title={Token preference optimization with self-calibrated visual-anchored rewards for hallucination mitigation},
  author={Gu, Jihao and Wang, Yingyao and Cao, Meng and Bu, Pi and Song, Jun and He, Yancheng and Li, Shilong and Zheng, Bo},
  journal={arXiv preprint arXiv:2412.14487},
  year={2024}
}

@article{wang2025perception,
  title={Perception-aware policy optimization for multimodal reasoning},
  author={Wang, Zhenhailong and Guo, Xuehang and Stoica, Sofia and Xu, Haiyang and Wang, Hongru and Ha, Hyeonjeong and Chen, Xiusi and Chen, Yangyi and Yan, Ming and Huang, Fei and others},
  journal={arXiv preprint arXiv:2507.06448},
  year={2025}
}

@article{huang2025spotlight,
  title={Spotlight on token perception for multimodal reinforcement learning},
  author={Huang, Siyuan and Qu, Xiaoye and Li, Yafu and Luo, Yun and He, Zefeng and Liu, Daizong and Cheng, Yu},
  journal={arXiv preprint arXiv:2510.09285},
  year={2025}
}

@article{li2026rethinking,
  title={Rethinking Token-Level Policy Optimization for Multimodal Chain-of-Thought},
  author={Li, Yunheng and Kuang, Hangyi and Zhang, Hengrui and Cao, Jiangxia and Liu, Zhaojie and Hou, Qibin and Cheng, Ming-Ming},
  journal={arXiv preprint arXiv:2603.22847},
  year={2026}
}

@article{liu2025noisyrollout,
  title={Noisyrollout: Reinforcing visual reasoning with data augmentation},
  author={Liu, Xiangyan and Ni, Jinjie and Wu, Zijian and Du, Chao and Dou, Longxu and Wang, Haonan and Pang, Tianyu and Shieh, Michael Qizhe},
  journal={arXiv preprint arXiv:2504.13055},
  year={2025}
}

@article{li2025self,
  title={Self-rewarding vision-language model via reasoning decomposition},
  author={Li, Zongxia and Yu, Wenhao and Huang, Chengsong and Liu, Rui and Liang, Zhenwen and Liu, Fuxiao and Che, Jingxi and Yu, Dian and Boyd-Graber, Jordan and Mi, Haitao and others},
  journal={arXiv preprint arXiv:2508.19652},
  year={2025}
}

@article{bai2025qwen3,
  title={Qwen3-vl technical report},
  author={Bai, Shuai and Cai, Yuxuan and Chen, Ruizhe and Chen, Keqin and Chen, Xionghui and Cheng, Zesen and Deng, Lianghao and Ding, Wei and Gao, Chang and Ge, Chunjiang and others},
  journal={arXiv preprint arXiv:2511.21631},
  year={2025}
}

@inproceedings{lu2021inter,
  title={Inter-gps: Interpretable geometry problem solving with formal language and symbolic reasoning},
  author={Lu, Pan and Gong, Ran and Jiang, Shibiao and Qiu, Liang and Huang, Siyuan and Liang, Xiaodan and Zhu, Song-Chun},
  booktitle={Proceedings of the 59th Annual Meeting of the Association for Computational Linguistics and the 11th International Joint Conference on Natural Language Processing (Volume 1: Long Papers)},
  pages={6774--6786},
  year={2021}
}

@article{wang2025vl,
  title={Vl-rethinker: Incentivizing self-reflection of vision-language models with reinforcement learning},
  author={Wang, Haozhe and Qu, Chao and Huang, Zuming and Chu, Wei and Lin, Fangzhen and Chen, Wenhu},
  journal={arXiv preprint arXiv:2504.08837},
  year={2025}
}

@article{qiao2407we,
  title={We-math: Does your large multimodal model achieve human-like mathematical reasoning?, 2024},
  author={Qiao, Runqi and Tan, Qiuna and Dong, Guanting and Wu, Minhui and Sun, Chong and Song, Xiaoshuai and GongQue, Zhuoma and Lei, Shanglin and Wei, Zhe and Zhang, Miaoxuan and others},
  journal={URL https://arxiv. org/abs/2407.01284}
}

@inproceedings{guan2024hallusionbench,
  title={Hallusionbench: an advanced diagnostic suite for entangled language hallucination and visual illusion in large vision-language models},
  author={Guan, Tianrui and Liu, Fuxiao and Wu, Xiyang and Xian, Ruiqi and Li, Zongxia and Liu, Xiaoyu and Wang, Xijun and Chen, Lichang and Huang, Furong and Yacoob, Yaser and others},
  booktitle={Proceedings of the IEEE/CVF conference on computer vision and pattern recognition},
  pages={14375--14385},
  year={2024}
}

@article{lu2023mathvista,
  title={Mathvista: Evaluating mathematical reasoning of foundation models in visual contexts},
  author={Lu, Pan and Bansal, Hritik and Xia, Tony and Liu, Jiacheng and Li, Chunyuan and Hajishirzi, Hannaneh and Cheng, Hao and Chang, Kai-Wei and Galley, Michel and Gao, Jianfeng},
  journal={arXiv preprint arXiv:2310.02255},
  year={2023}
}

@inproceedings{zhang2024mathverse,
  title={Mathverse: Does your multi-modal llm truly see the diagrams in visual math problems?},
  author={Zhang, Renrui and Jiang, Dongzhi and Zhang, Yichi and Lin, Haokun and Guo, Ziyu and Qiu, Pengshuo and Zhou, Aojun and Lu, Pan and Chang, Kai-Wei and Qiao, Yu and others},
  booktitle={European Conference on Computer Vision},
  pages={169--186},
  year={2024},
  organization={Springer}
}

@inproceedings{yue2024mmmu,
  title={Mmmu: A massive multi-discipline multimodal understanding and reasoning benchmark for expert agi},
  author={Yue, Xiang and Ni, Yuansheng and Zhang, Kai and Zheng, Tianyu and Liu, Ruoqi and Zhang, Ge and Stevens, Samuel and Jiang, Dongfu and Ren, Weiming and Sun, Yuxuan and others},
  booktitle={Proceedings of the IEEE/CVF conference on computer vision and pattern recognition},
  pages={9556--9567},
  year={2024}
}

@article{chen2024we,
  title={Are we on the right way for evaluating large vision-language models?},
  author={Chen, Lin and Li, Jinsong and Dong, Xiaoyi and Zhang, Pan and Zang, Yuhang and Chen, Zehui and Duan, Haodong and Wang, Jiaqi and Qiao, Yu and Lin, Dahua and others},
  journal={Advances in Neural Information Processing Systems},
  volume={37},
  pages={27056--27087},
  year={2024}
}

@inproceedings{kembhavi2016diagram,
  title={A diagram is worth a dozen images},
  author={Kembhavi, Aniruddha and Salvato, Mike and Kolve, Eric and Seo, Minjoon and Hajishirzi, Hannaneh and Farhadi, Ali},
  booktitle={European conference on computer vision},
  pages={235--251},
  year={2016},
  organization={Springer}
}

@article{liu2024ocrbench,
  title={Ocrbench: on the hidden mystery of ocr in large multimodal models},
  author={Liu, Yuliang and Li, Zhang and Huang, Mingxin and Yang, Biao and Yu, Wenwen and Li, Chunyuan and Yin, Xu-Cheng and Liu, Cheng-Lin and Jin, Lianwen and Bai, Xiang},
  journal={Science China Information Sciences},
  volume={67},
  number={12},
  pages={220102},
  year={2024},
  publisher={Springer}
}
\bibliographystyle{unsrtnat}
% \newpage
% \appendix
% \input{sections/A_appendix}
% \newpage
% \input{checklist}
\end{document}